\documentclass[letterpaper, 10 pt, conference]{ieeeconf}  

\IEEEoverridecommandlockouts                              

\usepackage{graphics} 
\usepackage{epsfig} 
\usepackage{amsmath} 
\usepackage{amssymb}  

\usepackage{times}
\usepackage{algorithmic}
\usepackage{algorithm}
\usepackage{epstopdf}
\usepackage{multirow}
\usepackage{booktabs}
\usepackage[font=small]{caption}
\usepackage{subfigure}
\usepackage{cite} 
\usepackage{lipsum}
\usepackage{xcolor}
\usepackage{hyperref} 
\usepackage{gensymb}
\usepackage{arydshln}
\usepackage{setspace}

\usepackage{verbatim}

\usepackage{amsthm}

\usepackage{amstext}
\usepackage{fancyhdr}

\fancypagestyle{preprint}{
  \fancyhf{}
  \fancyhead[L]{PREPRINT VERSION}
  
  \setlength{\topmargin}{-0.7in}
  \setlength{\headheight}{15pt}
  \setlength{\headsep}{25pt}
}

\title{\LARGE \bf
Learning-based Dynamic Robot-to-Human Handover
}

\author{Hyeonseong Kim$^{1}$, Chanwoo Kim$^{1}$, Matthew Pan$^{2}$, Kyungjae Lee$^{3}$ and Sungjoon Choi$^{1}\textsuperscript{*}$
\thanks{${^1}$Hyeonseong Kim, Chanwoo Kim and Sungjoon Choi are with the Department of Artificial Intelligence, Korea University, Seoul, Republic of Korea \texttt{\{hyeonseong-kim, chanwoo-kim, sungjoon-choi\}@korea.ac.kr}}
\thanks{${^2}$Matthew Pan is with the Department of Electrical and
Computer Engineering, Queens University, Kingston, Canada \texttt{\{matthew.pan\}@queensu.ca}}
\thanks{$^{^3}$Kyungjae Lee is with the Department of Statistics, Korea University, Seoul, Republic of Korea \texttt{\{kyungjae\_lee\}@korea.ac.kr}}
}
\begin{document}

\pagestyle{preprint}
\maketitle
\thispagestyle{preprint}

\begin{abstract}
This paper presents a novel learning-based approach to dynamic robot-to-human handover, addressing the challenges of delivering objects to a moving receiver. 
We hypothesize that dynamic handover, where the robot adjusts to the receiver's movements, results in more efficient and comfortable interaction compared to static handover, where the receiver is assumed to be stationary. 
To validate this, we developed a nonparametric method for generating continuous handover motion, conditioned on the receiver's movements, and trained the model using a dataset of 1,000 human-to-human handover demonstrations. 
We integrated preference learning for improved handover effectiveness and applied impedance control to ensure user safety and adaptiveness. 
The approach was evaluated in both simulation and real-world settings, with user studies demonstrating that dynamic handover significantly reduces handover time and improves user comfort compared to static methods.
Videos and demonstrations of our approach are available at \textbf{\url{https://zerotohero7886.github.io/dyn-r2h-handover/}}.
\end{abstract}

\section{Introduction}
As robots become more integrated into everyday environments, seamless interaction between humans and robots is crucial for effective collaboration.
One of the most fundamental tasks in this context is robot-to-human handover, where an object is transferred from a robot to a human in an intuitive and efficient manner.

While previous research \cite{rasch2018joint,aleotti2012comfortable,he2022go,strabala2013toward,moon2014meet,faibish2022human,newbury2022visualizing,8542485} has made progress in developing more intuitive and efficient handover systems, these approaches often assume that the human remains stationary, with the object being delivered to a fixed position. 
Although some studies \cite{meng2022fast, 8967614, mohammed2024presentation, yang2022model, christen2023learning, christen2024synh2r, wang2024genh2r} have investigated real-time hand tracking for smoother handovers, the receiver is still limited to a confined area during the process.
In real-world situations, handovers often occur in dynamic contexts, such as receiving a flyer while walking through a crowded street or taking a water bottle during a marathon. For a seamless handover, the robot must be able to adapt to the receiver's movements in real-time.

In this paper, we hypothesize that dynamic handover is not only more efficient but also feels more comfortable and natural for the human compared to static handover. 
To prove this, we develop a dynamic handover system that allows the receiver to move during the handover process, with the robot reactively adjusting its actions in real-time. 

A key challenge in developing a dynamic handover system is predicting and adapting to human movements while ensuring that the robot's actions remain both safe and natural. 
This requires accurately predicting the human's position to generate intuitive motions and controlling the robot to ensure smooth coordination with human actions. 
Ensuring safety in physical human-robot interaction (pHRI)\cite{de2008atlas} is particularly demanding, as the robot must balance responsiveness with the ability to manage unpredictable changes in human movement. 
One of the primary challenges in pHRI is optimizing impedance parameters~\cite{sharifi2021impedance, okada2023learning, lee2012human, yu2020estimation}, such as stiffness and damping, which are difficult to fine-tune due to their task-specific and interdependent nature. 
These complexities make dynamic handover a particularly intricate and demanding task.

To overcome these challenges, our dynamic handover system involves the policy learned from human data. 
While prior research has explored dynamic handovers~\cite{kupcsik2018learning}, it lacked the integration of real human motion data for real-time trajectory adaptation. 
The key distinction of our work is leveraging collected 1,000 human-to-human handover motion data for an intuitive handover process. 
Additionally, preference learning~\cite{Fürnkranz2011} was used to fine-tune jointly related control parameters, including impedance variables, from human feedback.

In summary, this paper introduces a dynamic handover system that incorporates data-driven human modeling, adaptive pose prediction, preference learning optimization, and compliant robot control. 
Quantitative evaluations show that our trajectory generation module outperforms baseline models in predicting the giver's pose based on receiver motions. 
By jointly optimizing the control parameters through preference learning, we improve the system's intuitiveness and responsiveness. 
Human-subject evaluations confirm that our approach outperforms static handover methods in both comfort and efficiency, highlighting the potential of dynamic handover systems for enhancing human-robot collaboration across various applications.

\section{Problem Formulation}
This research aims to develop a dynamic handover system and compare it with static handover in terms of time efficiency and user comfort.
In this study, dynamic handover refers to the scenario where the receiver is in motion, requiring the robot to adapt continuously to changes in pose. 
Static handover, by contrast, assumes the receiver remains stationary. 
The main challenge is enabling the dynamic handover system to adjust in real-time to the receiver's movements, which involves generating force targets for the robot to ensure smooth and compliant physical interactions.

\subsubsection{Input}
The primary input to the system is \textbf{the observed receiver's trajectory} $\boldsymbol{\xi}^\mathbf{\textbf{r}}_{\textbf{o}}$, which is a continuous stream of the human receiver until the current time $t_c$. 
In our study, we focus on the position of the base $x^{\text{r}}_{\text{o}} \in \mathbb{R}^2$.
The robot must process this real-time trajectory data to predict and respond to the receiver's future movements during the dynamic handover.

\subsubsection{Ouput}
The output of the system is the robot end-effector's \textbf{desired force} $\mathbf{F} \in \mathbb{R}^6$ for Cartesian impedance control~\cite{siciliano2008springer}. Since both the human and robot are in motion and physical interaction occurs during the handover, controlling the robot purely with position-based control can result in rigid and unnatural movements. To avoid this, the system generates force targets that guide the robot's behavior, enabling it to adapt compliantly to the receiver's forces.
\subsubsection{Core Challenges}
A key challenge in the dynamic handover system is \textbf{real-time robot trajectory generation}, requiring continuous adaptation to the receiver's movements. 
To address this, we leverage motion patterns observed in human-human interactions, which are naturally intuitive to humans.
Another significant challenge lies in implementing an impedance controller that enables \textbf{force-based tracking} of the generated robot trajectory. 
While flexibly responding to interactions with the receiver, the robot must continuously measure the forces exerted on its end-effector to ensure proper releasing timing. 

\section{Preliminaries}
\subsection{Cartesian Impedance Control}
In our system, we adopt Cartesian impedance control~\cite{siciliano2008springer} to regulate the interaction dynamics between the robot end-effector and the human in Cartesian space. The objective of impedance control is to make the robot's behavior compliant, as if governed by a virtual spring-damper system, allowing it to respond adaptively to external forces during tasks such as handover.

Assuming that $\mathbf{x} \in \mathbb{R}^6$ is the position and orientation of the end-effector in Cartesian space, we express the desired impedance behavior in Cartesian space as
\begin{equation}
    \mathbf{F} = M (\Delta\ddot{\mathbf{x}}) + B (\Delta\dot{\mathbf{x}}) + K (\Delta{\mathbf{x}}),
\end{equation}
where \( \Delta{\mathbf{x}} = \mathbf{x}_{\text{d}} - \mathbf{x} \), \( \Delta\dot{\mathbf{x}} = \dot{\mathbf{x}}_{\text{d}} - \dot{\mathbf{x}} \), and \( \Delta\ddot{\mathbf{x}} = \ddot{\mathbf{x}}_{\text{d}} - \ddot{\mathbf{x}} \) represent the position, velocity, and acceleration errors between the desired pose \( \mathbf{x}_{\text{d}} \) and the current pose \( \mathbf{x} \) of the end-effector.
The matrices \( M \), \( B \), and \( K \) are the desired inertia, damping, and stiffness, respectively. 
The output force \( \mathbf{F} \) generated by this control law is used to guide the robot's end-effector, ensuring to adapt compliantly to external forces and interactions with the human, such as during object handover.

In our study, inertia  $M$ is set to a constant value due to the system limitation, and focuses on the tuning of \( B \) and \( K \). Additionally, we assume that Coriolis forces, gravity, and other nonlinear effects in the robotic system are compensated by the underlying controller.

\subsection{Preference Learning}
In this work, we employ the widely adopted Gaussian process-based preference learning algorithm, CoSpar \cite{tucker2020preference}. The goal is to enhance human satisfaction in dynamic handover systems by jointly optimizing handover parameters that significantly impact human-robot interactions.

Let $\mathcal{A} \subseteq \mathbb{R}^d$ represent a finite set of actions with $A = |\mathcal{A}|$. 
The dataset $D = \{x_k \succ x_k'\}_{k=1}^N$ contains $N$ preference observations, where each pair $(x_k, x_k')$ indicates that the human prefers action $x_k$ over $x_k'$.
We assume a latent utility function $f(x)$ for each action $x \in \mathcal{A}$, where the utility vector is given by $\mathbf{f} := [f(x_1), f(x_2), \dots, f(x_A)]^\top$.
In \cite{chu2005preference, tucker2020preference}, the likelihood for a preference $x_k \succ x_k'$ is obtained as
\begin{align}\label{likelihood}
P(x_k \succ x_k') = \Phi \left( \frac{f(x_k) - f(x_k')}{\sqrt{2}\sigma} \right),
\end{align}
where $\Phi(\cdot)$ is the cumulative distribution function of the standard normal distribution. 
Then, the full likelihood is expressed as
\begin{align}\label{full_likelihood}
P(D \mid f) = \prod_{k=1}^N \Phi \left( \frac{f(x_k) - f(x_k')}{\sqrt{2}\sigma} \right).
\end{align}
The posterior distribution $P(f \mid D)$ can then be approximated using Laplace approximation~\cite{chu2005preference}.

\section{Methodology}
\begin{figure*}[!t] 
    \centering
    \includegraphics[width=1.9\columnwidth]{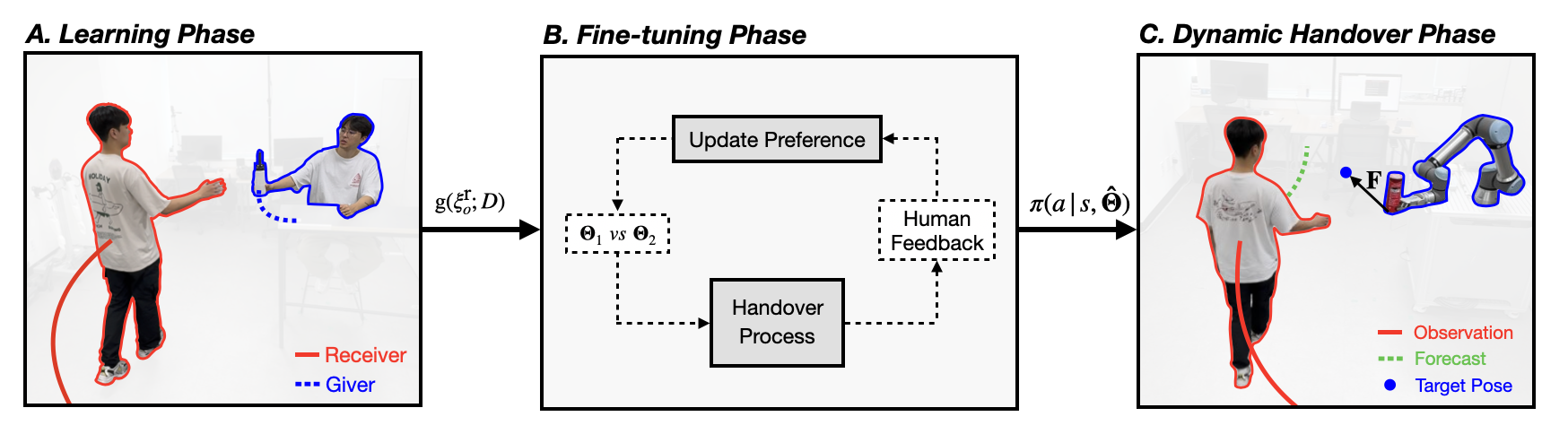}
    \caption{\textbf{Overview of dynamic handover system.} In \textit{A. Learning Phase}, the robot trajectory generation module $\textit{g}(\xi^\text{r}_o;D)$,  which predicts the robot trajectory on the observed receiver trajectory $\xi^{\text{r}}_{o}$, is trained using the human-to-human handover dataset $D$. In \textit{B. Fine-tuning Phase}, preference learning adjusts the handover parameters $\boldsymbol{\Theta}$ including impedance variables, based on human feedback from real-world handover trials. Finally, in \textit{C. Dynamic Handover Phase}, the system utilizes the learned $\textit{g}$ and ${\hat{\boldsymbol{\Theta}}}$ to select desired force $\mathbf{F}$ at end-effector in real-time, ensuring smooth and responsive handover performance.}
    \label{fig:overview}
    \vspace{-10pt}
\end{figure*}

The proposed dynamic handover system consists of three development phases: learning, fine-tuning, and dynamic handover, shown in Fig.~\ref{fig:overview}.
In the learning phase, human-to-human handover data is collected to build a robot trajectory generation model, predicting trajectories based on the receiver's movements. 
The fine-tuning phase uses preference learning to optimize key parameters, such as impedance variables and release force thresholds. 
Finally, in the dynamic handover phase, the system is deployed on the robot in real-time using the learned model and parameters.

\subsection{Learning Phase}
In the learning phase, we first collect a human-to-human handover dataset. 
Inspired by IAAL~\cite{choi2018nonparametric}, we then develop and train a nonparametric robot trajectory generation model, denoted by $\textit{g}(\xi^\text{r}_o;D)$, where $\xi^\text{r}_o$ is the observed receiver trajectory and $D$ is the collected dataset.
Our proposed model measures the similarities between various trajectories and predicts the robot's pose sequence from the present into the future based on these similarities.

\subsubsection{Data Collection}
First, we define the receiver pose \( \mathbf{x}^\text{r} \in \mathbb{R}^2 \) as the receiver base \( xy \)-pose, and the giver pose \( \mathbf{x}^\text{g} \in \mathbb{R}^3 \) as the giver wrist \( xyz \)-pose. 
Each trajectory, \( \xi^\text{r} \) for the receiver and \( \xi^\text{g} \) for the giver, consists of a sequence of these poses, represented as \( \{\mathbf{x}^\text{r}_k\}_{t=1}^{L_k} \) and \( \{\mathbf{x}^\text{g}_k\}_{t=1}^{L_k} \), respectively, where \( L_k \) is the trajectory length. 
We collected a dataset \( D = \{\xi_{k}^\text{r}, \xi_{k}^\text{g}\}_{k=1}^{N} \) from 1,000 human-to-human handovers ($N=1000$), where the receiver moves along a path while the giver hands over an object.

\begin{figure}[!t] %
    \centering
    \includegraphics[width=1.0\columnwidth]{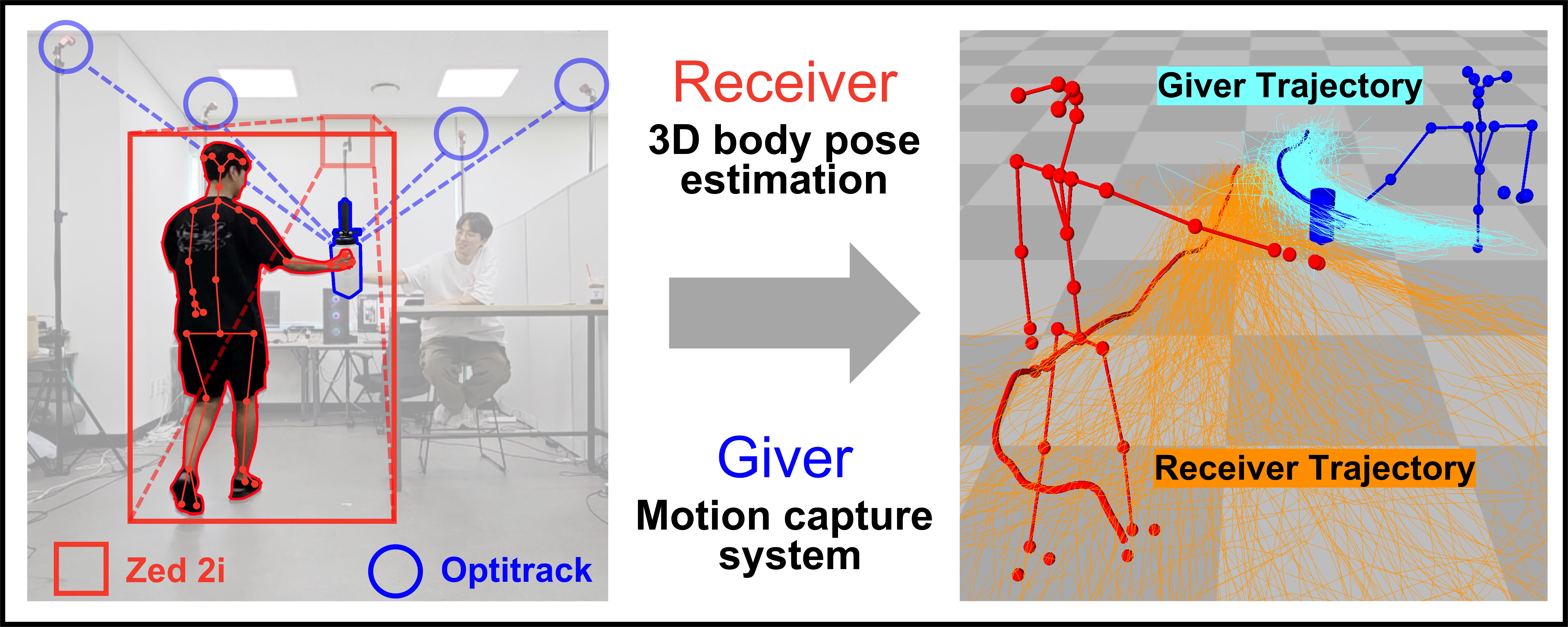}
    \caption{\textbf{Human-Human data collection.} 1,000 pairs of receiver-giver trajectory are obtained using the 3D body pose estimation and motion capture system.}
    \label{fig:dataset_collection}
    \vspace{-10pt}
\end{figure}

As demonstrated in Fig.~\ref{fig:dataset_collection}, ZED 2i Stereo Camera was positioned overhead to ensure successive tracking of the receiver, with the 3D body pose estimated using ZED SDK 4.0~\cite{stereolabs_zed_sdk}. 
Additionally, 11 OptiTrack Flex 13 cameras with Motive 3.0~\cite{optitrack_motive} captured precise motion data of the object during the handover. 
Due to the relatively high noise and estimation error in 3D human body pose estimation, we use the object pose from the motion capture system as the giver's wrist pose.
The global base is fixed at the receiver xy pose and z is fixed at the floor.

\subsubsection{Trajectory Similarity Measure}
We define observed receiver pose until current time $t_c$ as $ \mathbf{x}^\text{r}_o \in \mathbb{R}^2 $ and $\xi^\text{r}_o$ =  $\{\mathbf{x}^\text{r}_{o,t}\}_{t=1}^{t_c}$  and $\xi^\text{r}_k$ as a trajectory in collected receiver trajectories.
The flow function \( f: \mathbf{x} \rightarrow \dot{\mathbf{x}} \) maps the pose to the velocity, which is modeled using a Sparse Pseudo-input Gaussian Process (SPGP)~\cite{snelson2005sparse}. 
The mean \( \hat{\mu} \) corresponds to the estimated velocity vector $\dot{\mathbf{x}}$ given $\mathbf{x}$ as an input. 
The trajectory distance \( d(\xi^\text{r}_o; \xi^\text{r}_k) \) is defined as
\begin{equation}
d(\xi^\text{r}_o; \xi^{r}_k) = \frac{1}{t_c} \sum_{t=1}^{t_c} \left( \kappa \cdot d_{\text{cos}} (\dot{\mathbf{x}}^\text{r}_{o,t}, \hat{\mu}_k(\mathbf{x}^\text{r}_{o,t})) + \hat{\sigma}_k^2(\mathbf{x}^\text{r}_{o,t}) \right)
\label{eq:d}
\end{equation}
where $d_{\text{cos}}(\dot{\mathbf{x}}_a, \dot{\mathbf{x}}_b) = 1 -(\dot{\mathbf{x}}_a^T \dot{\mathbf{x}}_b / \|\dot{\mathbf{x}}_a\|_2 \|\dot{\mathbf{x}}_b\|_2) $, \(\hat{\mu}_k(\cdot)\) and \(\hat{\sigma}_k^2(\cdot)\) are the mean and variance produced by the flow function $f(\cdot;\xi^{\text{r}}_k)$, modeled by SPGP for $\xi^{\text{r}}_k$. 
In our SPGP, we define an inducing ratio between 0 and 1, representing the proportion of actual data points used as inducing points. 
The selection of these inducing points considers that \( \xi\) is time-series data, ensuring a uniform distribution across time.
\( \kappa \) is a hyperparameter that controls the relative contribution of the cosine distance \( d_{\text{cos}} \) and the variance term \( \hat{\sigma}_k^2 \). 
Intuitively, \( d(\xi^\text{r}_o; \xi^\text{r}_k) \) represents the measurement of the distance of $\xi^\text{r}_o$ about $\xi^\text{r}_k$ which considers both the direction of the velocity (temporal distance) and the position difference (spatial distance).
Then, the trajectory similarity of \( \xi^\text{r}_o\) about \(\xi^\text{r}_k \) can be defined as
\begin{equation}
sim_k(\xi^\text{r}_o) = e^{-d(\xi^\text{r}_o; \xi^\text{r}_k)}.
\label{eq:sim}
\end{equation}
\subsubsection{Robot Trajectory Generation}
Utilizing the introduced similarity measure, we can generate the predicted giver trajectory \( \xi^{\text{pred}} \) from \( \{ \xi^{\text{r}}_{k}, \xi^{\text{g}}_{k} \}_{k=1}^N \), which will be treated as the desired robot trajectory. 
First, compute the similarity of \( \xi^{\text{r}}_{\text{o}} \) with every \( \xi^{\text{r}} \in D \), and select the most similar \( K \) trajectory indices with \eqref{eq:sim}. 
Let $\mathcal{S}$ denote this set of selected $K$ indices. 
For each $k \in \mathcal{S}$, we can find the time index 
\( t_k \), on which the receiver pose is closest to the current observed receiver pose, and extract the corresponding giver trajectory. 
The predicted trajectory \( \xi^{\text{pred}} \) is computed using \eqref{eq:pred}, where \( T \) is the minimum of \( L_k \) and \( w_k \) represents the normalized $sim_k$ in the range \([0,1]\), ensuring that their sum equals 1. 

\begin{equation}
\xi^\text{pred} = \sum_{k \in \mathcal{S}} w_k \cdot \xi^{\text{g}}_{k, t_{k}:T}
\label{eq:pred}
\end{equation}

\subsection{Fine-tuning Phase} 
In the fine-tuning phase, we optimize the handover parameters using a preference-based learning algorithm, CoSpar~\cite{tucker2020preference}. The parameters, denoted by \( \boldsymbol{\Theta} = \{K, B, t_f, f_r\} \), include the stiffness and damping variables, a forecasting time for compensating time lag, and a release threshold. Each of these parameters is interrelated, and they are jointly optimized based on human preference feedback collected through the handover process to ensure smooth and adaptive handover performance. Below are the explanations of each objective parameter.
\subsubsection{Stiffness and Damping (\( K \) and \( B \))}
The stiffness \( K \) and damping \( B \) parameters work together to regulate the robot's force response according to position and velocity error during handover. To simplify the process, the rotational components of both K and B are fixed, and only the position-related terms, treated as variables, are optimized. We can balance the tracking performance and adaptiveness of handover interaction by fine-tuning \( K \) and \( B \).
\subsubsection{Forecasting Time (\( t_f \))}
One of the challenges in impedance control is dealing with inherent time lag due to the system's dynamic nature. Additionally, since we use camera-based human tracking, the maximum frequency at which we can update the target pose is limited by the camera's frame rate, exacerbating the time lag. To compensate for this, we introduce the forecasting time parameter \( t_f \), which allows the robot to track a future target pose \(\mathbf{x}_{d,t+t_f} \) instead of the current target pose \(\mathbf{x}_{d,t} \). This is possible because our estimation module predicts the future trajectory $\xi^\text{pred}$ from \eqref{eq:sim}. Importantly, \( t_f \) is closely related to the stiffness \( K \) and damping \( B \) parameters, as all three need to be balanced to ensure smooth tracking without excessive lag or instability.
\subsubsection{Releasing Threshold (\( f_r \))}
The releasing threshold $f_r$ defines the force required for the gripper to release the object. If set too low, the robot may drop the object prematurely; if too high, the human must apply excessive force, leading to discomfort. This parameter is critical for user comfort as it directly impacts the handover experience.

In summary, the fine-tuning phase optimizes these key handover parameters through the CoSpar algorithm, allowing the robot to adapt dynamically to the human's behavior during the handover. The interdependency between \( K \), \( B \), and \( t_f \) ensures that the robot is responsive and anticipates future movements, while \( f_r \) is carefully calibrated to maximize user comfort during the release phase.

\subsection{Dynamic Handover Phase}
The system executes real-time handover in the dynamic handover phase using the learned parameters, \( \hat{\boldsymbol{\Theta}} \), obtained from the fine-tuning phase. During this phase, the robot dynamically adapts its actions to the receiver's movements based on $K$, $B$, and $f_r$. While the target pose is given by the robot trajectory generation model and $f_r$, the rotation target is fixed except for yaw, so that the robot gripper maintains a horizontal orientation during the handover. In addition to Cartesian impedance control, we apply a temporal ensemble technique, inspired by \cite{zhao2023learning}, to refine the target pose trajectory. This approach allows the system to generate a smooth desired force $\mathbf{F}$ in real-time, ensuring compliant and adaptive handover behavior. This is possible because our robot trajectory generation module is capable of continuously forecasting future target trajectories. During the handover process, the gripper is opened if external force is above $f_r$ and the handover process is terminated.

\section{Experiments}
We conduct both simulation and human-subject experiments to assess the advantages of our proposed robot trajectory generation module and dynamic handover system.
In Sec.~\ref{simulation}, we provide a quantitative comparison of our trajectory generation module against baseline methods.
In Sec.~\ref{userstudy}, we describe the user study design, including the real-world implementation with a robot, and demonstrate the practical effectiveness of proposed dynamic handover system in terms of user comfort and handover processing time.

\subsection{Simulation Experiment}\label{simulation}
Before demonstrating the advantages of the dynamic handover system, we first validate the prediction performance of the proposed robot trajectory generation module in terms of root mean squared (RMS) errors between the predicted and the ground truth giver trajectories from the collected data. 
Sample efficiency is measured in the relation between training data size and RMS error, and inference time to see how fast the trajectory generation module can infer.
The trajectory generation module receives the base pose $\mathbf{x}^\text{r} \in \mathbb{R}^2$ (excluding height) as an input and predicts the giver pose $\mathbf{x}^\text{g} \in \mathbb{R}^3$.

\subsubsection{Baselines} 
We compared the proposed trajectory generation module with three existing approaches: behavior cloning with a multi-layer perceptron (BC-MLP)~\cite{jang2022bc, zhang2018deep}, action chunking with transformer (ACT)~\cite{zhao2023learning} and intention aware apprenticeship learning (IAAL)~\cite{choi2018nonparametric}.
BC-MLP is a widely used supervised learning technique for visuomotor tasks, extracting actions from images processed by a convolutional neural network. 
ACT, a more recent behavior cloning method, leverages a transformer architecture to predict action sequences and has demonstrated superior performance in real-world tasks~\cite{zhao2023learning,fu2024mobile}. 
Finally, IAAL introduces a nonparametric motion flow model that uses Gaussian Process (GP) to compute trajectory similarities. 
Our trajectory generation module shares similarities with IAAL but employs SPGP to enable real-time control.

\subsubsection{Implementation Details}
The collected dataset consists of 900 in-distribution (ID) trajectories, where the receiver moves directly toward the giver to receive the object, and 100 out-of-distribution (OOD) trajectories, in which the receiver either wanders or pauses during the handover process.
In ~\ref{fig:dynamic_handover}, we provide snapshots of the dynamic handover in ID and OOD scenarios with detailed explanations.
The results for the proposed method in Tab.~\ref{tab:robustness} and Fig.~\ref{fig:sample_eff} use an inducing rate of \(0.4\).

\begin{table}[!b]
    \centering
    \begin{tabular}{lcc}
    \toprule
    \multirow{2}{*}{\textbf{Method}} & \multicolumn{2}{c}{\textbf{RMS Error (mm)}} \\ 
    \cmidrule(lr){2-3}
    & \textbf{ID} & \textbf{OOD} \\ 
    \midrule
      BC-MLP               & 152.27 $\pm$ 289.11 & 166.50 $\pm$ 311.60 \\ 
      ACT                  & 67.57  $\pm$ 5.06   & 71.04  $\pm$ 7.14   \\
      IAAL                 & \textbf{55.10 $\pm$ 2.28} & \textbf{61.82 $\pm$ 4.82} \\ 
      Ours                 & \textbf{55.13  $\pm$ 2.48}   & \textbf{61.83  $\pm$ 5.32} \\
    \bottomrule
    \end{tabular}
    \caption{\textbf{RMS error.} The values in each cell represent the RMS error (mean $\pm$ std) between the predicted and ground truth giver trajectories.}
    \label{tab:robustness}
\end{table}

\subsubsection{Simulation Results}
The simulation results presented below are acquired by partitioning the dataset into 10 subsets using a stratified k-fold algorithm. 
A fixed seed was used for testing to ensure consistency in the results. 
Tab. \ref{tab:robustness} summarizes the RMS error between the predicted and ground truth giver trajectories for both the proposed method and the baseline approaches.
The results indicate that the nonparametric trajectory generation modules, especially IAAL, achieve the lowest errors in robot trajectory prediction.
IAAL's strong performance can be attributed to its use of GP to represent motion flows, whereas our method incorporates SPGP to account for inference time efficiency. 
Despite this, it is clear that nonparametric methods offer advantages in robot trajectory prediction, likely due to their ability to directly compare entire trajectories.
Furthermore, nonparametric approaches appear to be more robust, maintaining their predictive accuracy even when the receiver exhibits unusual behaviors.

\begin{figure}[!t]
    \centering
    \includegraphics[width=0.8\columnwidth]{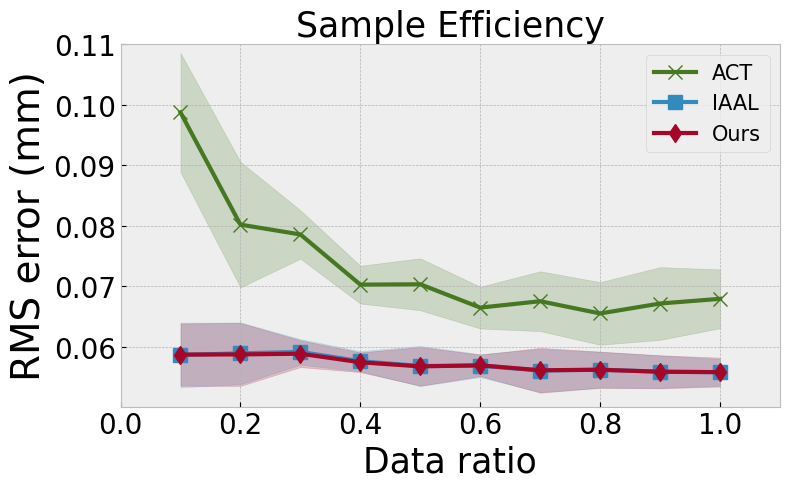}
    \caption{\textbf{Sample efficiency.} The values within each marker represent the RMS error (mean $\pm$ std) of the trajectory generation modules for different training data sizes.}
    \label{fig:sample_eff}
    \vspace{-10pt}
\end{figure}

The nonparametric methods also demonstrate benefits to sample efficiency.
As shown in Fig.~\ref{fig:sample_eff}, the RMS error increases as the dataset size decreases. 
However, while nonparametric approaches exhibit only a slight increase in RMS error with reduced data, ACT experiences a more significant rise in error as the data ratio decreases.
For clarity, BC-MLP results are omitted but follow a similar trend to ACT, with a higher error increase at smaller data sizes. 
These observations underscore the robustness of nonparametric methods, particularly when handling limited datasets.

\begin{figure}[!t]
    \centering
    \includegraphics[width=0.8\columnwidth]{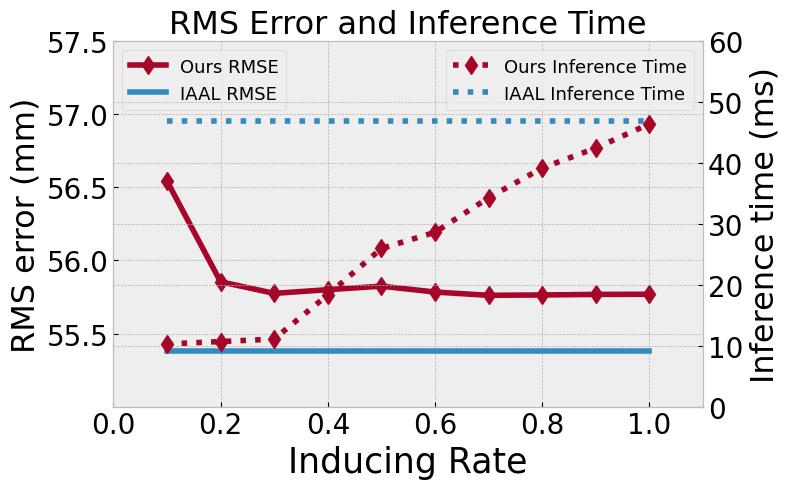}
    \caption{\textbf{Relationship between RMS error and inference time.} The values within each marker represent the RMS error and inference time for IAAL and varying inducing rates for the proposed method.}
    \label{fig:tradeoff}
    \vspace{-10pt}
\end{figure}

While IAAL demonstrates superior performance in robot pose prediction, Fig.~\ref{fig:tradeoff} shows that utilizing SPGP achieves comparable RMS error to IAAL while significantly improving inference time. 
This aspect allows our trajectory generation module to enable real-time robot control while maintaining a decent robot trajectory prediction, making it a more practical solution for a time-sensitive dynamic handover system.

\subsection{Human Subject User Study}\label{userstudy}
In this section, we conducted a human-subject user study to evaluate the advantages of the dynamic handover system in human-robot interaction compared to static handover. 
Additionally, we examined the impact of preference learning, which jointly optimizes the parameters of the dynamic handover system.
The study focuses on three key questions:
\begin{itemize}
    \item How do users perceive comfort in dynamic versus static handover?
    \item How efficient is dynamic handover compared to static handover in terms of handover processing time?
    \item How does the joint optimization of handover parameters related to compliance control and handover, through preference learning, influence the user experience during object reception?
\end{itemize}

\begin{figure*}[!t]
    \centering
    \subfigure[User Study Instruction]{
    \includegraphics[width=0.4\columnwidth,height=0.7\columnwidth]{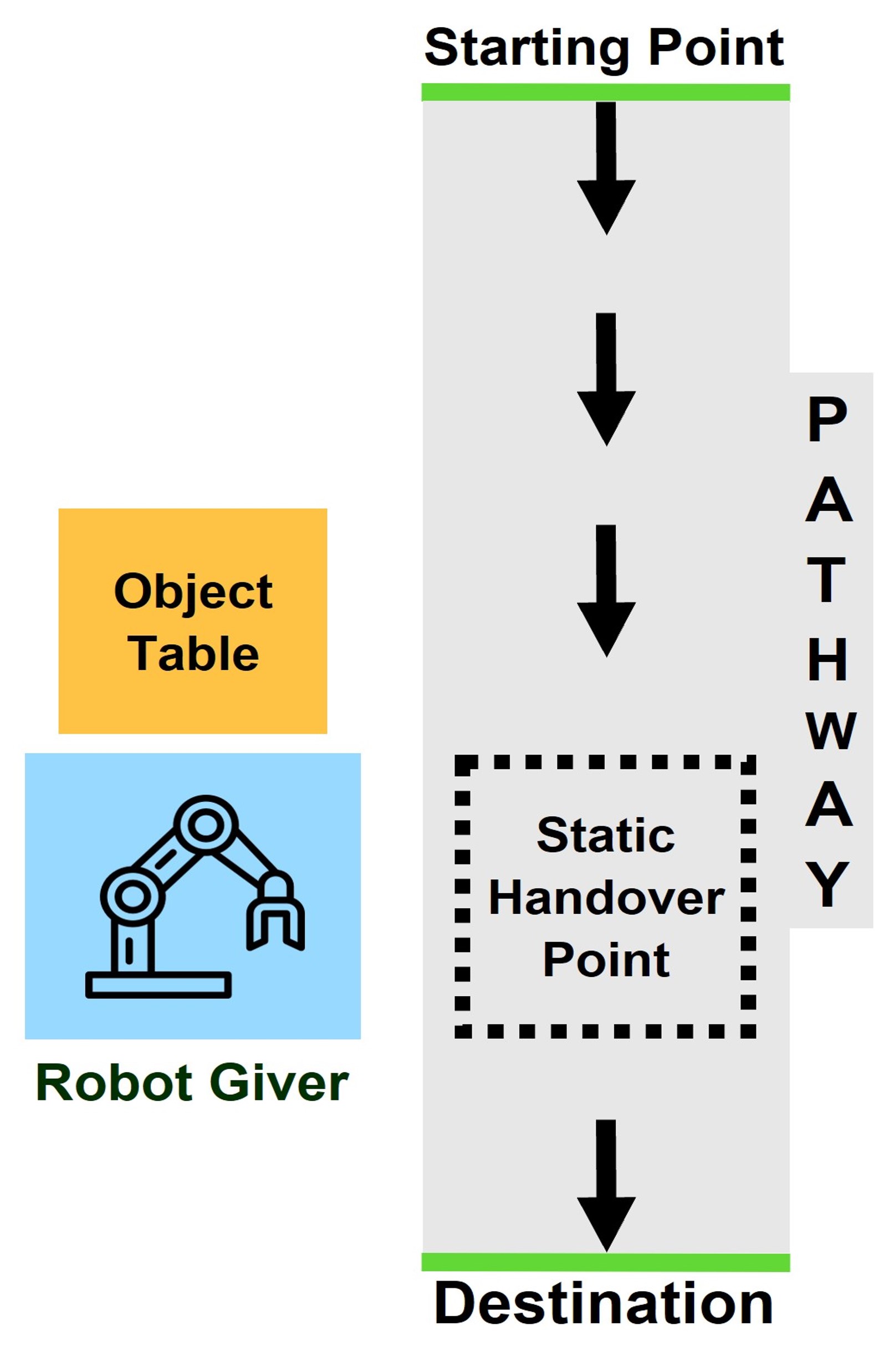}
    \label{fig:instruction}}
    \hspace{0.02\textwidth} 
    \subfigure[Dynamic Handover Demonstrations]{
    \includegraphics[width=1.45\columnwidth,height=0.7\columnwidth]{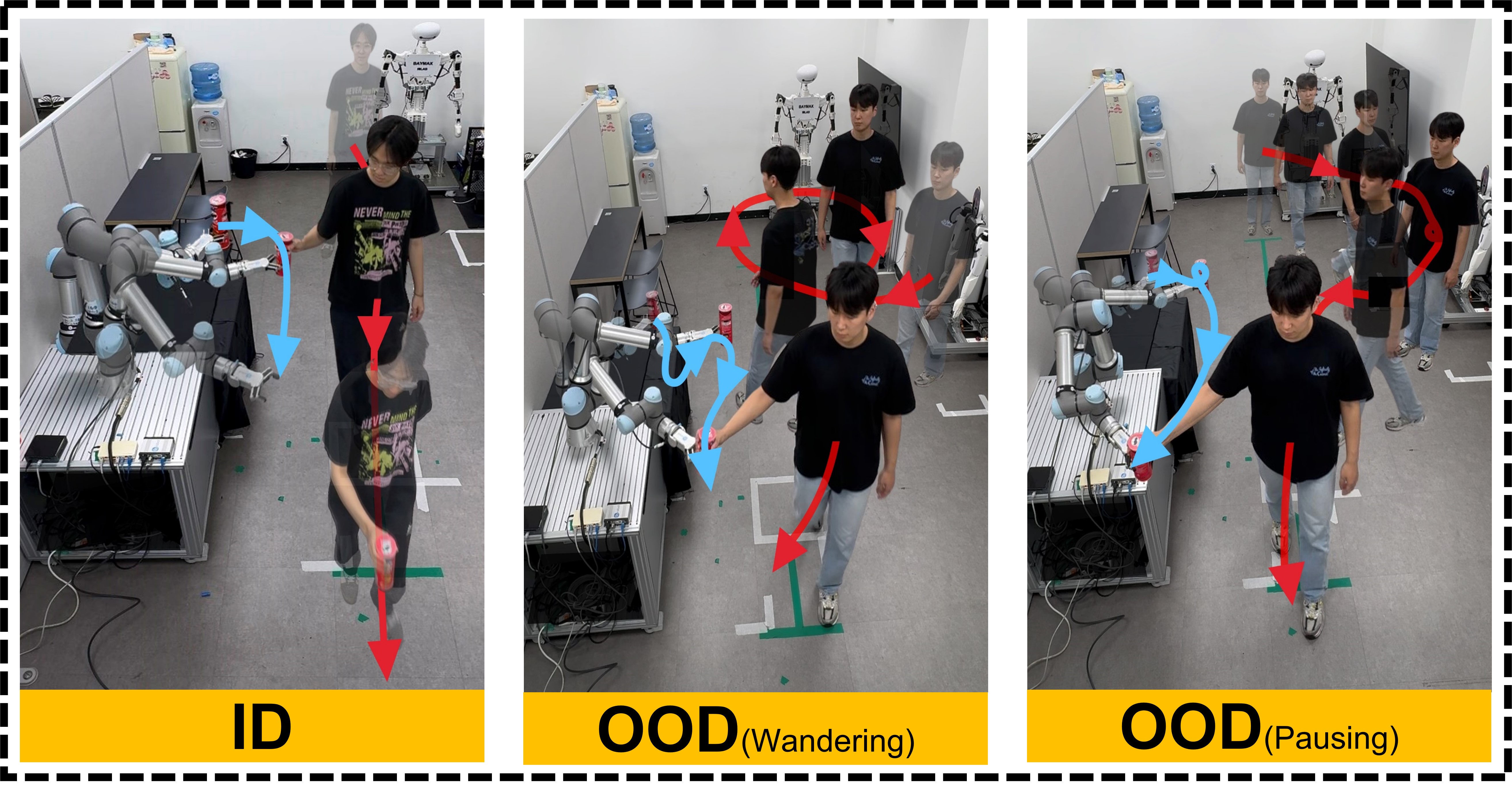}
    \label{fig:dynamic_handover}}
    \caption{\textbf{Real world dynamic handover demonstrations.} Following the instruction in Fig.~\ref{fig:instruction}, real-world dynamic handover demonstrations in Fig.~\ref{fig:dynamic_handover} are implemented with the UR5e in a real-world setting. 
    In the dynamic handover scenarios, the left figure represents the ID scenario, the middle figure illustrates the OOD scenario with humans moving unpredictably, and the right figure shows the OOD scenario where humans alternate between stopping and moving.}
    \vspace{-10pt}
\end{figure*}

\subsubsection{Real-World Demonstrations}\label{real_demo}
For real-world deployment, a ZED 2i stereo camera with the Body Tracking API was employed to capture the receiver's base pose at 30 Hz.
The robot base was placed at the global base pose to align the robot coordinate system with the giver pose.
Control parameters were jointly optimized using CoSpar over 20 iterations, within the ranges \( K = [80,140] \), \( B = [10,20] \), \( t_f = [0,1] \), and \( f_r = [5,20] \), resulting in the final learned parameters \( \hat{\boldsymbol{\Theta}} = \{K, B, t_f, f_r\} =  [114.3, 17.1, 0.14, 7.1] \). 
Additionally, the temporal ensemble technique is applied with a chunk size of 30 (1 second) to smooth the target pose output.
The UR5e manipulator~\cite{ur5e_robot} tracked the target pose using impedance control with the learned \( K \) and \( B \) while the control loop operated at 200 Hz, utilizing $f_r$ for predictive tracking. 
As shown in Fig.~\ref{fig:dynamic_handover}, the proposed pipelines effectively handled dynamic handovers, safely and naturally adjusting to the receiver's movements, even in response to unexpected behavior like wandering or pausing.

\subsubsection{Dynamic \& Static Handover}\label{dynamic_static}
In this section, we assessed the effectiveness of the dynamic handover system compared to static handover by evaluating both comfort level and average handover processing time. 

As illustrated in Fig.~\ref{fig:instruction}, the eight participants (age: 20s-30s, 6 males and 2 females) were instructed to receive the object from the robot and then reach the destination on the opposite side.
For the static handover, we follow the approach in \cite{meng2022fast}, where the receiver enters a predefined area, and the robot delivers the object upon detecting the receiver's right hand.
To ensure stable recognition, the robot initiates movement after a 2-second delay following hand detection. 
This stabilization delay is excluded from the subsequent analysis of results for the average processing time.
For the dynamic handover in Fig.~\ref{fig:dynamic_handover}, they were allowed to receive the object at any point along the passageway.

We hypothesize that (\textbf{H1}) \textit{Participants will feel more comfortable during dynamic handover}, and (\textbf{H2}) \textit{Dynamic handover will result in a shorter average processing time than static handover.} 
To validate these hypotheses, a 5-point Likert scale \textit{(1: not at all, 5: very much)} was designed to assess the participants' comfort level in response to the statement (\textbf{Q1}) \textit{"The handover process felt comfortable"}.
Additionally, we measured the total time from the start to the moment participants received the object to evaluate the average handover processing time.
Each handover processing time was measured from when the participants left the starting point to when the robot's gripper released the object in both dynamic and static handover scenarios.
To ensure fairness, the delay for recognition stability in the static handover was subtracted.
The volunteers completed five iterations of both dynamic and static handovers.
After each iteration, participants answered \textbf{Q1} to evaluate their comfort level.

\begin{figure}
    \centering
    \subfigure[Measured Comfort Level]{
    \includegraphics[width=0.45\columnwidth]{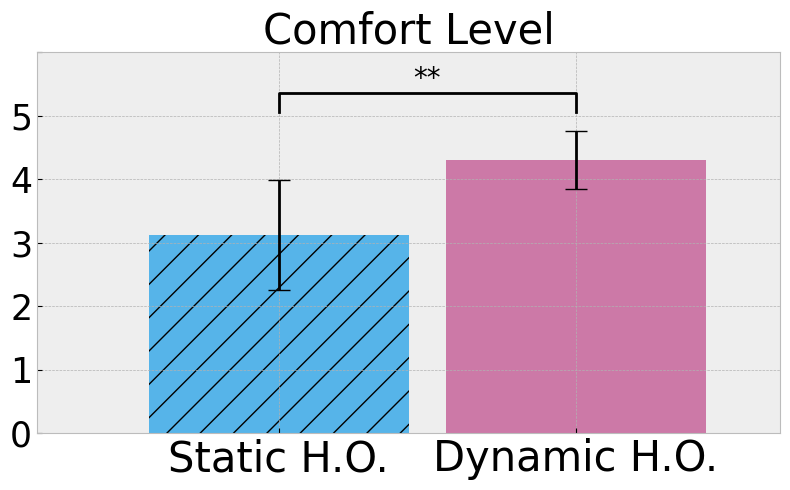}
    \label{fig:comfort_level}}
    \vspace{0.02\textwidth}
    \subfigure[Handover Processing Time]{
    \includegraphics[width=0.45\columnwidth]{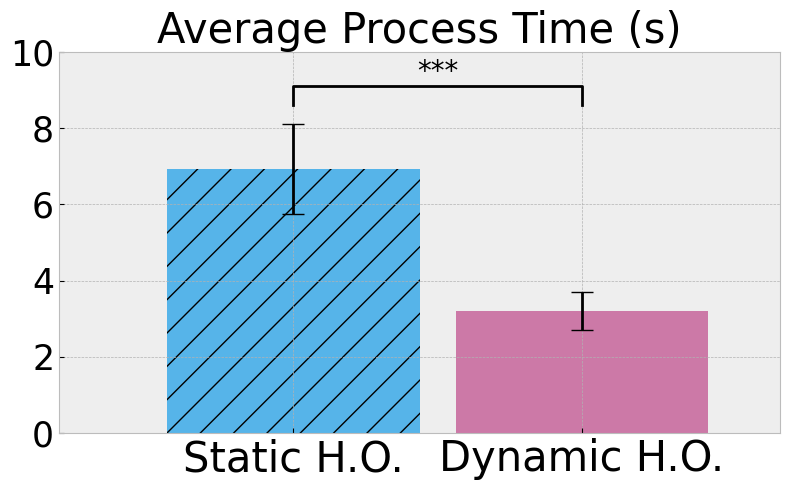}
    \label{fig:process_time}}
    \caption{User study results about comparing dynamic and static handover.}
    \vspace{-10pt}
\end{figure}

As shown in Fig.~\ref{fig:comfort_level}, participants rated dynamic handover significantly more comfortable (\( M = 4.3, SD = 0.45 \)) than static handover (\( M = 3.1, SD = 0.87 \)), supporting \textbf{H1} (\( p < 0.01 \), Cohen's \( d = 0.79 \)).
Fig.~\ref{fig:process_time} demonstrates that it also reduced handover time (\( M = 3.2s, SD = 0.50 \)) compared to static handover (\( M = 6.9s, SD = 1.2 \)), confirming \textbf{H2} (\( p < 0.001 \), Cohen's \( d = 1.92 \)).
Cohen's \( d \)~\cite{lakens2013calculating} is a standardized effect size measure for mean differences.
To prevent statistical inflation, individual trial averages were used. 

\subsubsection{Preference Learning Optimization}
We also examined the efficacy of preference learning optimization within the dynamic handover system. 
Our hypothesis (\textbf{H3}) was that \textit{Participants would be more satisfied with the dynamic handover system when using jointly optimized handover parameters compared to unoptimized ones.}
An A/B test was conducted with the volunteers each experiencing five iterations of both the optimized and unoptimized systems, after which they selected the more satisfactory option.
For the unoptimized system, handover parameters were randomly selected from the same feasible ranges used for training. 
$72.5\%$ of responses ($29$ out of $40$) indicated a preference for the optimized dynamic handover system over the unoptimized one. 
A two-tailed binomial test validates the effectiveness of joint parameter optimization, with statistical significance at $p<1e-02$ with a Cohen's $h$ of $0.47$ (interpretable as a medium effect size), supporting \textbf{H3}.

\subsection{Limitations}
Our preference learning approach assumes a generalized optimal solution but does not fully account for individual body dynamics. 
In particular, differences in user height can affect z-axis adjustments, and walking speed variations influence forecasting time. 
These factors suggest the need for a more personalized handover system.

\section{Conclusion}
This paper presents a novel approach to the problem of dynamic robot-to-human handover, demonstrating that adapting the robot's movements to the receiver's motion significantly improves both the efficiency and comfort of the handover process. 
Our nonparametric motion generation method, combined with preference learning and impedance control, enables the robot to deliver objects in a natural and intuitive way to the user while ensuring safety and adaptability.
The results of our experiments, conducted in both simulated and real-world environments, indicate that dynamic handover offers substantial benefits over static handover methods.

\vspace{-0.3cm}

\bibliographystyle{IEEEtran}
\bibliography{references}

\end{document}